\documentclass[journal]{IEEEtran}

\usepackage{xcolor,soul,framed} 

\colorlet{shadecolor}{yellow}
\usepackage[pdftex]{graphicx}
\graphicspath{{../pdf/}{../jpeg/}}
\DeclareGraphicsExtensions{.pdf,.jpeg,.png}

\usepackage[cmex10]{amsmath}
\usepackage{amsfonts}
\usepackage{array}
\usepackage{mdwmath}
\usepackage{mdwtab}
\usepackage{eqparbox}
\usepackage{url}
\usepackage{bm}
\usepackage{algorithm}
\usepackage{algorithmic}

\begin{document}
\bstctlcite{IEEEexample:BSTcontrol}
    \title{Modelling Networked Dynamical System by Temporal Graph Neural ODE with Irregularly Partial Observed Time-series Data}
  \author{Mengbang Zou, Weisi Guo \\
  \thanks{Mengbang Zou (corresponding author, emial:M.zou@cranfield.ac.uk)and Weisi Guo are with Cranfield University, Cranfield, MK43 0AL, U.K.}}
\maketitle

\begin{abstract}
Modeling the evolution of system with time-series data is a challenging and critical task in a wide range of fields, especially when the time-series data is regularly sampled and partially observable. Some methods have been proposed to estimate the hidden dynamics between intervals like Neural ODE or Exponential decay dynamic function and combine with RNN to estimate the evolution. However, it is difficult for these methods to capture the spatial and temporal dependencies existing within graph-structured time-series data and take full advantage of the available relational information to impute missing data and predict the future states. Besides, traditional RNN-based methods leverage shared RNN cell to update the hidden state which does not capture the impact of various intervals and missing state information on the reliability of estimating the hidden state. To solve this problem, in this paper, we propose a method embedding Graph Neural ODE with reliability and time-aware mechanism which can capture the spatial and temporal dependencies in irregularly sampled and partially observable time-series data to reconstruct the dynamics. Also, a loss function is designed considering the reliability of the augment data from the above proposed method to make further prediction. The proposed method has been validated in experiments of different networked dynamical systems. 
\end{abstract}


%
\IEEEpeerreviewmaketitle


\section{Introduction}
\IEEEPARstart{M}{athematical} models are fundamental for us to describe and understand the evolution of a system. However, in complex networked systems, data is abundant, while the physical laws and governing equations to model the interactions between components that co-evolve in time remain elusive \cite{brunton2022data}. How to model the time evolution of networked dynamical systems from time-series data is challenging and critical in a wide range of fields such as traffic prediction of base stations in communication networks \cite{wang2022spatial}, learning the evolution of dynamics in particle-based systems \cite{bishnoi2022enhancing}, traffic flow prediction in transportation system \cite{ma2021short}. Normally, we need to obtain complete time-series data with uniform intervals to model the system. However, time series data with non-uniform intervals often happen as well as the loss of information of components. 

\subsection{Review on regular time-series data with missing information}
For example, in sensor networks or the Internet of Things, faulty sensors and network failures are widespread phenomena that cause disruptions in the data acquisition process. For incomplete time-series data, in which, at a certain time step, missing data appears at some of the channels of the resulting multivariate time series. Several methods have been proposed to impute missing values in time series, such as imputation methods based on the k-nearest neighbours \cite{beretta2016nearest}, and matrix factorisation approximation methods \cite{cichocki2009fast}. Among different imputation methods, approaches based on deep learning have attracted a lot of attention \cite{cini2021filling, yoon2018estimating, yoon2018gain}. While classic imputation methods can be used to fill the missing values of the feature matrix, none of them can capture the underlying graph structure. Due to the ability to reserve the graph-structured nature of the data by encoding the underlying graph-structured data using topological relationships among the nodes of the graph, imputation methods based on GNN have been proposed. In \cite{cini2021filling}  a novel graph neural network architecture has been introduced, named GRIN, which aims at reconstructing missing data in the different channels of a multivariate time series by learning spatio-temporal representations through message passing. \cite{rossi2022unreasonable} presented a general approach for handling missing features in graph machine learning applications which is based
on minimization of the Dirichlet energy and leads to a diffusion-type differential equation on the graph.

\subsection{Review on irregular sampling time-series data}
In the above imputation methods, the intervals between observations in time-series data are constant. Time-series data with non-uniform intervals happens in many applications. For example, many real-world tasks for autonomous vehicles or robots need to integrate input from a variety of sensors with different sampling frequency \cite{neil2016phased, weerakody2021review}. In this situation, these imputation methods are not efficient any more. This is because most of the above imputation methods are based on recurrent neural networks (RNNs) framework when dealing with time series data. However, traditional RNN is ankward to deal with irregular time series data. A standard trick is to divide the timeline into uniform intervals and use a constant or undefined hidden states between intervals. Such prepocessing may destroy the information and lead to inaccurate modelling of the system. A better way solve this problem is to construct a continuous-time model with a latent state defined at all times. Che et al., proposed define RNNs with continuous dynamics given by a simple exponential decay between observations \cite{che2018recurrent}. An elegant method has been proposed in \cite{rubanova2019latent}, where the continuous dynamics between observations is estimated by a neural network as in Neural ODEs \cite{chen2018neural}. This model is called ODE-RNN, which can handle arbitrary time gaps between observations. Due to the power in dealing with graph-structured data, graph neural networks (GNN) have been combined with Neural ODEs to predict trajectories of dynamical systems with interacting components\cite{poli2019graph, luo2023hope, luo2024care}.

According to our survey, when modelling the evolution of networked systems with time-series data, most of the current research considers either regular sampling time-series data with missing information or irregular sampling time-series data. In this paper, we consider a more challengable task, which aims to model the evolution of the networked system by irregular sampling time-series data with partial observable information. To solve this problem, we propose a framework which combines an impute network and a prediction network. The impute part is a temporal Graph Neural ODE consisting of a Graph Neural ODE (GNODE) and a Graph Gate Recurrent Unit (GGRU). The Graph Neural ODE is applied to approximate the spatial and temporal evolution of hidden states between observations by a hidden continuous dynamics function estimated by a graph neural network. With the previous hidden state and the current observable state, Graph Convolutional Gate Recurrent Unit can be used to update the hidden state. However, traditional RNN-based methods
leverage shared RNN cell to update the hidden state which does not capture the impact of various intervals and missing state information on the reliability of estimating the hidden state, i.e., current observed state with less missing information and smaller time interval from last observation is more reliable to update the hidden state. To solve this problem, in this paper, we propose a reliability and time-aware mechanism which can capture the impact of various intervals and missing state information.

Hidden states are updated at each observable time step based on the ground-truth data, the imputation within observed time series is accurate. However, this impute network is not accurate in prediction because the framework based on RNN (GRU, LSTM) is trained for one-step ahead prediction which causes the accumulation of estimation error step by step. To solve this problem, we train another GNODE by the time-series data generated by the impute network. This prediction network is very easy to train compared with the impute network because no need to estimate and update step by step. Data generated by the impute network consists of the ground-truth data from observation and generative data. The generative data close to the observable data is more accurate and is more important for training. Therefore, each sample's weight of the data can be adjusted according to its data quality and each sample's weight is introduced to loss function for training. Here, we design an exponential decay function to calculate the weight of each sample. This enables the sample of higher quality to play a more important role in training the prediction network.

\subsection{Novelty and Contribution}
Modelling a dynamics system and predicting future states with irregularly partial observed time-series data is a challengable task. The contribution of this paper is that we propose a framework which consists of an impute network and a prediction network to model the evolution of networked system by irregular sampling and partial observable time-series data. The impute network is based on temporal Graph Neural ODE consisting of Graph Neural ODE and Graph Gate Recurrent Unit with reliability and time-aware mechanism. Unlike RNN-based method with sharing RNN cell to update hidden states, the proposed method which can capture the impact of various intervals and missing state
information as well as the spatial and temporal dependencies, enabling  accurately impute temporal and spatial data. The prediction network can make prediction by learning from the imputation data. Since the quality of the imputation data generated by the impute network is heterogeneous, an exponential decay function is designed to adjust the weight of the data to calculate the loss function. This enables the sample of higher quality to play a more important role in training the prediction network.

\section{Methods}

\begin{figure*}[ht]
    \centering
    \resizebox*{18cm}{!}{\includegraphics{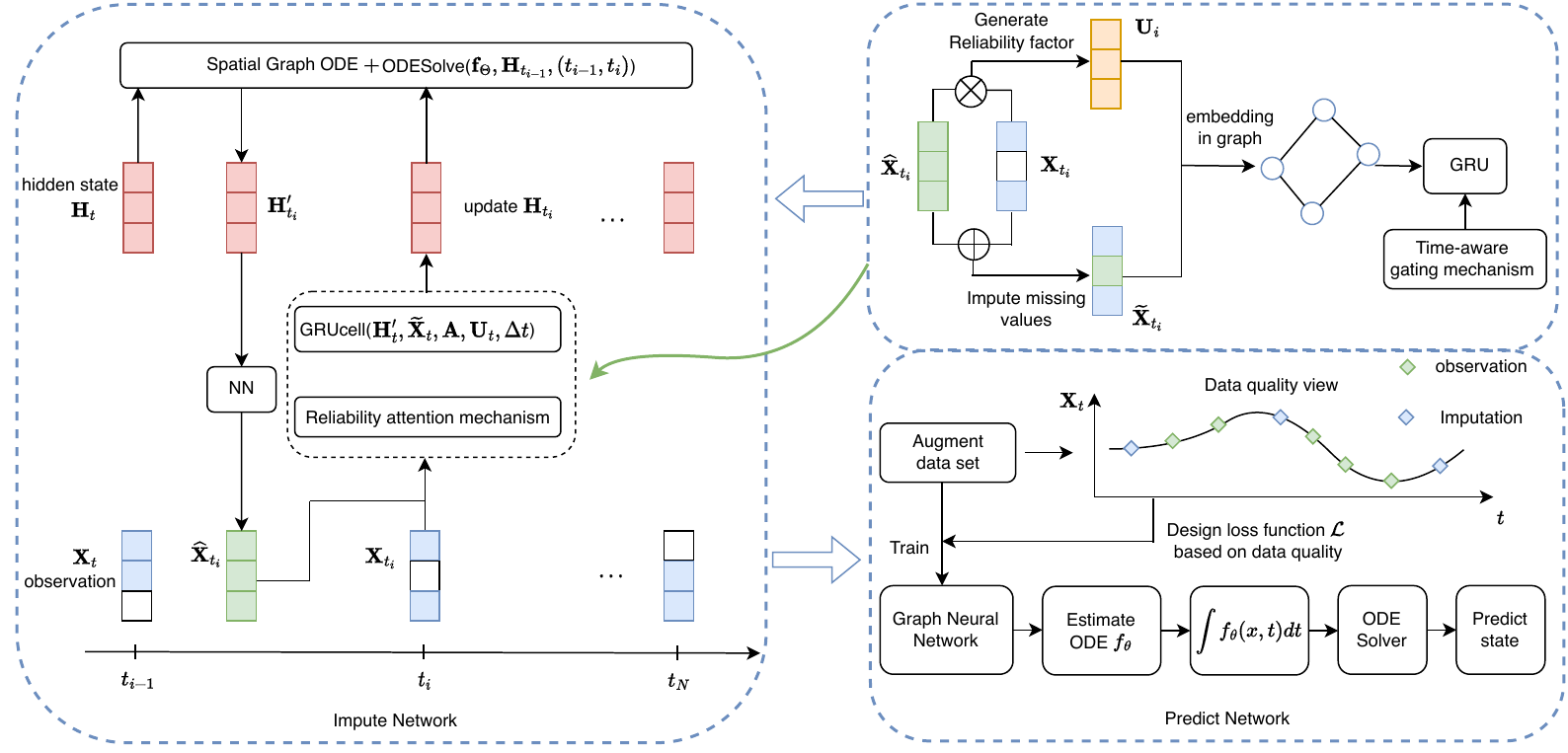}}
    \caption{The structure of GNODE-GGRU to impute data. The hidden state between observations is estimated by Graph neural ODE and then updated by Graph GRU. The impute data can be used to make prediction in predict network.}
    \label{fig: irregular_data}
\end{figure*}

\begin{figure}[ht]
    \centering
    \resizebox*{9cm}{!}{\includegraphics{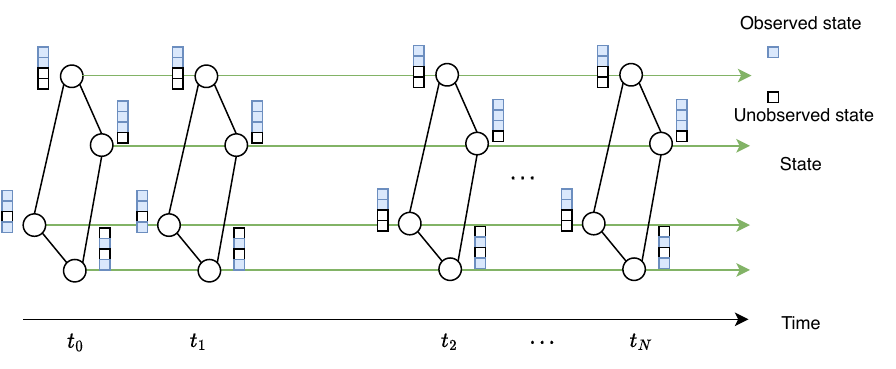}}
    \caption{Irregular sampling time-series graph-structured data with unobserved states. The blue one represents the observed state and the white one is unobserved.}
    \label{fig: missing_data}
\end{figure}

\begin{figure*}[ht]
    \centering
    \resizebox*{16cm}{!}{\includegraphics{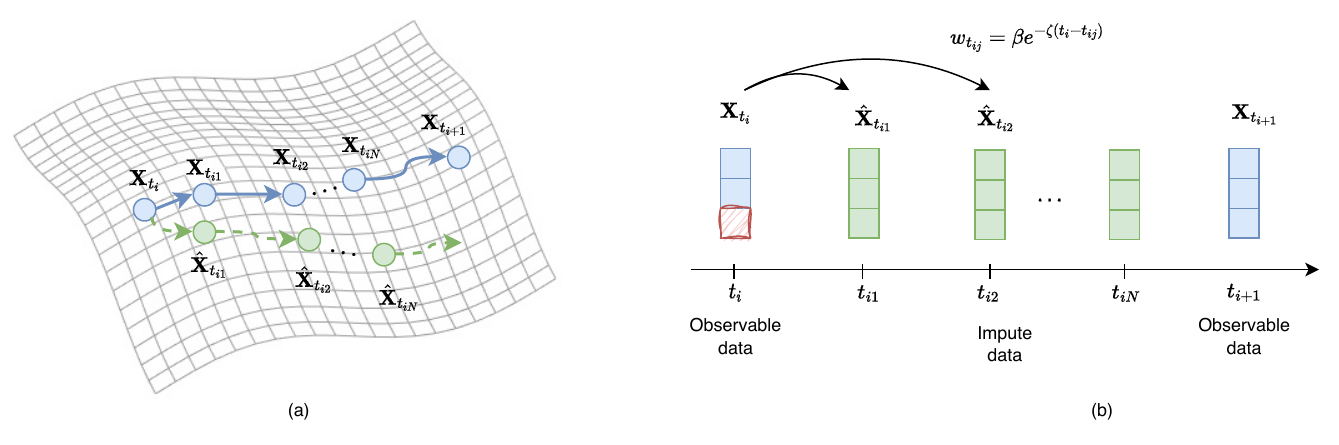}}
    \caption{Blue one is the observable state and the red one is the missing information. Green one is the impute state by GNODE-GGRU. Weight of each term in loss function is determined by $ w_{t_{ij}} = \beta e^{-\zeta (t_i-t_{ij})}$.}
    \label{fig: loss_function}
\end{figure*}

\begin{figure*}[ht]
    \centering
    \resizebox*{16cm}{!}{\includegraphics{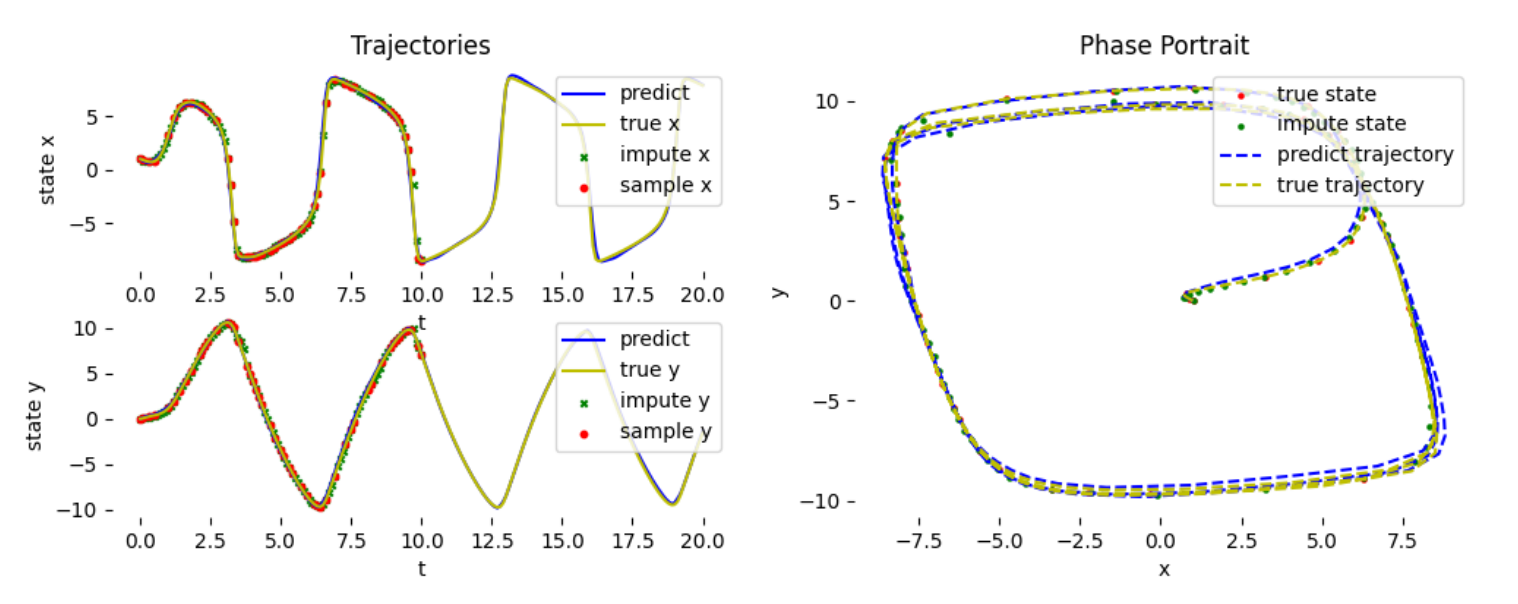}}
    \caption{Modelling the trajectories of the system according to the temporal irregular and spatial missing data by the proposed method in this paper. The system contains 8 nodes and each node has a 2-dimensional dynamic function. The left figure shows trajectories of node 1 on $x$ plane and $y$ plane. The right figure is the phase portrait of node 1.}
    \label{fig: predict}
\end{figure*}

\subsection{Model of networked system}
In a networked system of $N$ nodes with nonlinear dynamics, the dynamics of each node can be described by an ODE as 
\begin{equation}\label{equ: coupling_dynamics}
    \dot{\bm{x}}^i=f(\bm{x}^i)+\sum_{j \neq i}^{N}a_{ij}g(\bm{x}^i, \bm{x}^j),
\end{equation}
where $\bm{x}^i$ is the state of node $i$, $\bm{x}^i \in \mathbb{R}^d$ and $d$ is the dimension of the state. $f(\cdot)$ is the self-dynamics of each node, $g(\cdot)$ is the coupling dynamics between node $i$ and node $j$ and $a_{ij}$ is the element of the adjacency matrix $\bm{A}$ in which $a_{ij}=1$ if node $i$ is connected with node $j$. Sometimes, we do not know the accurate dynamic function $f(\cdot)$ and $g(\cdot)$ of the system. Instead, we can only observe the state of each node $\bm{x}_t^i$ at time step $t$. $\bm{X}_t \in \mathbb{R}^{N \times d}=[\bm{x}_t^1, \bm{x}_t^2, \cdots, \bm{x}_t^N]$ is the matrix of state at time step $t$. Notice that we consider the irregular sampling data of the system, so the interval between different time steps $\Delta t_i = t_{i+1}-t_{i}$ is not a constant value. 
    RNN is a widely used method in time series data with the update function
    
\begin{equation}\label{equ: rnn}
    \bm{H}_{t_i}={\rm RNNCell}(\bm{H}_{t_{i-1}}, \bm{X}_{t_i}),
\end{equation}
where $\bm{H}_{t_i}$ is the hidden state at time step $t_i$. A problem in handling the irregular time-series data is how to define the hidden state between observations, since $\Delta t_i = t_{i+1}-t_{i}$ is not a constant value. If we directly use the traditional RNN in equation~(\ref{equ: rnn}) to deal with irregular time series data, it means that the hidden state between observations is constant, which may destroy the information of the hidden state. One way to solve this problem is to use a Neural ODE to estimate the hidden state between observations $t_i$ and $t_{i-1}$, and update at observation time step $t_i$.

\subsection{Spatial Graph Neural ODE}
Neural ODEs are a family of continunous-time models which defines a hidden state $\bm{H}_{t_i}$ as a solution to ODE initial-value problem $\frac{d\bm{H}(t)}{dt}=\bm{F}(\bm{H}(t),\Theta, t)$. The function $\bm{F}$ specifies the dynamics of the hidden state, using a neural network with parameters $\Theta$. The hidden state $\bm{H}_{t_i}$ can be calculated at any time step using a numerical ODE solver:

\begin{equation}\label{equ: ODESOLVE}
    \bm{H}_{t_i} = {\rm ODESolve} (\bm{F}, \bm{H}_{t_0},  (t_0, t_i))= \bm{H}_{t_0}+\int_{t_0}^{t_i}\bm{F}(\bm{H}(t), \Theta, t)dt
\end{equation}

In Neural ODE, the function $\bm{F}$ is approximated by a neural network (NN). The state between observations can be defined by the solution of an ODE: $\bm{H}'_{t_i}={\rm ODESolve} (\bm{F}, \bm{H}_{t_{i-1}}, (t_{i-1}, t_i))$. Then $\bm{H}_{t_i}$ can be updated by equation (\ref{equ: rnn}). The state $\bm{X}_{t_i}$ at any time between two observations is predicted by the corresponding $\bm{H}_i$. While we are free to choose any kind of neural network to estimate $\bm{F}$, ignoring the potential physical structure of the system may cause the neural network to be accurate within the training date set but fail in testing data.

Since we aim to model the networked system, the dynamics of the system has the form of equation~(\ref{equ: coupling_dynamics}). State of each component is decided by its self-dynamic function and its neighbours' states. Therefore, it is natural to consider using a kind of graph neural network to estimate the function $\bm{F}$. The dynamics of the hidden state could be written as 
\begin{equation}
    \dot{\bm{h}}^i = \Phi_{\rm self}(\bm{h}^i) + \sum^{N}_{j\neq i}a_{ij}\Phi_{\rm coup}(\bm{h}^i, \bm{h}^j),
\end{equation}
where $\bm{h}^i \in $ is the hidden state of node $i \in \mathbb{R}^{d_1}$. Here, we use a specific GNN to estimate the dynamics of the hidden state. $
\Phi_{\rm self}$ can be estimated by a MLP. Generally, the coupling dynamics can be estimated by a $\kappa$ layers GNN which incorporates $\kappa$ adjacent matrix $\bm{A}$ to aggregate information through walks of length $\kappa$. With the increase of layers in GNN, each node can aggregate information from further neighbour nodes. However, GNN suffers from the oversmoothing problem, a phenomenon where all node features in a deep GNN converge to the same constant value as the number of hidden layers is increased \cite{rusch2022graph}. To ensure $\Phi_{\rm coup}$ can capture the nonlinear relationship between neighbour nodes and avoid powers of adjacency matrix $\bm{A}^{\kappa}$, $\Phi_{\rm coup}$ is estimated by
$\Phi_{\rm coup}(\bm{h}^i, \bm{h}^j) = \phi(\bm{h}^i||\bm{h}^j)$, where $\phi$ is a multi-layers neural networks, the symbol $||$ denotes concatenation operator.
\begin{equation} \label{equ: coup_nn}
    \dot{\bm{h}}^i= \gamma(
    \bm{h}^i
    )+\bigoplus\limits_{j \in \mathcal{N}_i}\phi(\bm{h}^i_{k-1} || \bm{h}^j_{k-1}),
\end{equation}
where $\bigoplus$ denotes the function sum and $\gamma$ is a multi-layers neural network.The advantage of equation~(\ref{equ: coup_nn}) is that it is computationally efficient because it has a clear structure to estimate the self-dynamics and coupling-dynamics to avoid unnecessary aggregation of information from other nodes. In addition, it avoids oversmoothing problem by only using one layer adjacency matrix.

The problem is that the complete information of $\bm{X}_t$ is not always available at any observation. To impute the presence of missing values, we consider a binary mask $\bm{M}_t \in \{0, 1\}^{N*d}$ where each row $\bm{m}_t^i$ indicates whether the node features of $\bm{x}^i$ at time step $t$ are available. For example, if $m^{i,j}_t=0$, the information of $x^{i,j}_t$ is unavailable. Otherwise, if $m^{i,j}_t=1$, the information of $x^{i,j}_t$ is available. 

In the first step, we need to make the initial imputation of $\widetilde{\bm{X}}_{t_0}$ by $\widetilde{\bm{X}}_{t_0} = \bm{M}_{t_0} \odot \bm{X}_{t_0}+\overline{\bm{M}}_{t_0} \odot \sigma(\bm{H}_{0}\bm{V}_0+\bm{b}_0)$, where $\bm{V}_0$ and $\bm{B}_0$ are learnable matrix. Generally, the initial hidden state $\bm{H}_0$ is unknown, and we can simply assume that $\bm{H}_0$ is a $0$ matrix. The later hidden state between observations can be estimated by equation~(\ref{equ: ODESOLVE}) and updated by equation~(\ref{equ: rnn}). The state $\bm{X}_{t_i}$ with missing values can be imputed by

\begin{equation}\label{equ: pred}
    \widetilde{\bm{X}}_{t_{i}} = \bm{M}_{t_{i}} \odot \bm{X}_{t_{i}}+\overline{\bm{M}}_{t_i} \odot \widehat{\bm{X}}_{t_i},
\end{equation}
where $\widehat{\bm{X}}_{t_i}=\sigma(\bm{H}_{t_{i-1}}\bm{V}_s+\bm{B}_s)$ is reconstructed by the proposed method.

\subsection{Temporal Graph Neural Network}
RNN is a general method to work with time series data, but face the short-term memory problem due to the gradient vanishing problem. Here, we use Gate Recurrent Unit (GRU), which is designed to overcome the short-term memory problem, to learn the hidden state and predict the next state. The main structure of a traditional GRU is
\begin{equation}\label{equ: GRU}
 \begin{aligned}
    \bm{R}_{t_i} &= \sigma(\bm{W}_{xr}\widetilde{\bm{X}}_{t_i} + \bm{W}_{hr}\bm{H}_{t_{i-1}})\\
    \bm{Z}_{t_i} &= \sigma(\bm{W}_{xz}\widetilde{\bm{X}}_{t_i} + \bm{W}_{hz}\bm{H}_{t_{i-1}})\\
    \bm{C}_{t_i} &= {\rm tanh}(\bm{W}_{xc}\widetilde{\bm{X}}_{t_i} + \bm{W}_{hc}(\bm{R}_{t_i} \odot \bm{H}_{t_{i-1}}))\\
    \bm{H}_{t_i} &=\bm{Z}_{t_i} \odot \bm{H}_{t_{i-1}}+(1-\bm{Z}_{t_i}) \odot\bm{C}_{t_i},
 \end{aligned}
\end{equation}
where symbol $\odot$ denotes the Hadamard product, $\bm{W}_{xr}, \bm{W}_{hr}, \bm{W}_{xz}, \bm{W}_{hz}, \bm{W}_{xc}, \bm{W}_{hc}$ are learnable weight matrices.  

Since the time series data is graph-structured, we implement temporal graph convolutional network instead of the traditional GRU.  For given $\bm{h}^i_k$, i.e. the feature of node $i$ at $k$ layer, $\bm{h}^i_k$ is updated by a MPNN defined as 
\begin{equation}
    {\rm MPNN}(\bm{h}^i_k) = \gamma_k(\bm{h}^i_{k-1}, \bigoplus\limits_{j \in \mathcal{N}_i}\phi_k(\bm{h}^i_{k-1}, \bm{h}^j_{k-1})),
\end{equation}
$\gamma_k$ and $\phi_k$ denote differentiable functions such as single-layer or multi-layers neural network. For computational efficiency, the MPNN is specific as

\begin{equation}
    {\rm MPNN}(\bm{H}_i, \bm{W}_i) = \bm{W}_{i1}\bm{H}_i+\bm{W}_{i2}\bm{AH}_i
\end{equation}
Equation~(\ref{equ: GRU}) is replaced by 
\begin{equation} \label{equ: GGRU}
\begin{aligned}
     \bm{R}_{t_i} &= \sigma({\rm MPNN}(\widetilde{\bm{X}}_{t_i}||\bm{H}_{t_{i-1}}), \bm{W}_{ri}) \\
     \bm{Z}_{t_i} &=  \sigma({\rm MPNN}(\widetilde{\bm{X}}_{t_i}||\bm{H}_{t_{i-1}}), \bm{W}_{zi})\\
    \bm{C}_{t_i} &= {\rm tanh}({\rm MPNN}(\widetilde{\bm{X}}_{t_i}||\bm{R}_{t_i} \odot \bm{H}_{t_{i-1}}))
\end{aligned}
\end{equation}

According to equation~(\ref{equ: pred}), $\widetilde{\bm{X}}_{t_{i}}$ consists of $\bm{X}_{t_i}$ and $\widehat{\bm{X}}_{t_i}$. $\bm{X}_{t_i}$ with less missing information enables $\widetilde{\bm{X}}_{t_{i}}$ more reliable. To quantify the reliability of state $\widetilde{\bm {X}}_{t_i}$ at any time step, one method is to combine $\bm{M}_{t_i}$ with $\widetilde{\bm{X}}_{t_{i}}$ as a feature vector \cite{che2018recurrent}. This method is straightforward to identify the importance of observed states but ignores the fact that prediction states at different time steps actually have different reliability. For example, if $\bm{M}_{t_{i}} \odot \bm{X}_{t_{i}}$ is close to $\bm{M}_{t_i} \odot \widehat{\bm{X}}_{t_i}$, then it is reasonable to think that it is reliable to use $\overline{\bm{M}}_{t_i} \odot \widehat{\bm{X}}_{t_i}$ to fill the missing values in $\overline{\bm{M}}_{t_i} \odot \bm{X}_{t_i}$. Otherwise, it is not an accurate estimation of $\overline{\bm{M}}_{t_i} \odot \bm{X}_{t_i}$. Therefore, we use a reliability factor matrix $\bm{U}_{t_i}$ to quantify the reliability of $\widetilde{\bm{X}}_{t_i}$ at any time step. At $t_i$, element $u_{ij}$ in $\bm{U}_{t_i}$ is calculated by
\begin{equation}
    u_{ij} = \left\{
    \begin{array}{lr}
         1 , &{\rm if } \quad m_{ij}=1; \\
    \frac{1}{1+|\alpha|}, &{\rm otherwise},
    \end{array}  
    \right.
\end{equation}
where $ \alpha = \frac{\sum_{i=1}^{N}\sum_{j=1}^dm_t^{i,j}(\widehat{x}_t^{i,j}-x_t^{i,j})^2}{\bm{1}^{\top}\bm{M}_{t_i}\bm{1}}$. Then equation~(\ref{equ: GGRU}) is rewritten as 
\begin{equation}
    \begin{aligned}
        \bm{R}_{t_i} &= \sigma({\rm MPNN}_2(\widetilde{\bm{X}}_{t_i}||\bm{U}_{t_i}||\bm{H}_{t_{i-1}}), \bm{W}_{ri}) \\
     \bm{Z}_{t_i} &=  \sigma({\rm MPNN}_2(\widetilde{\bm{X}}_{t_i}||\bm{U}_{t_i}||\bm{H}_{t_{i-1}}), \bm{W}_{zi})\\
    \bm{C}_{t_i} &= {\rm tanh}({\rm MPNN}_2(\widetilde{\bm{X}}_{t_i}||\bm{U}_{t_i} ||\bm{R}_{t_i} \odot \bm{H}_{t_{i-1}}))
    \end{aligned}
\end{equation}

Notice that time intervals between observations are not the same. As the increase of the time interval, it is more difficult for equation~(\ref{equ: ODESOLVE}) to estimate the hidden dynamics between intervals. This is because the hidden states between intervals follow complex trajectories but are determined by the last hidden state. The error in the last hidden state may be enlarged with time. For example, $\dot{x}(t) = x, x(t_0)=x_0$, the solution of $x(t)=x_0e^t$. If there exists an error $\Delta x_0$ in $x_0$, then the error in $x(t)$ is $\Delta x_0 e^t$. Also, the numerical methods like Euler methods, Runge-Kutta methods to calculate the Neural ODE will accumulate errors with the time interval. This means that the hidden state estimated in with a smaller interval is more reliable. However, ODE-RNN in equation~(\ref{equ: rnn}) leverages a simple shared RNN cell to update the hidden state at any time step, which does not capture the impact of various intervals on the reliability.  Therefore, it is reasonable to consider the factor $\Delta t$ in forget gate $\bm{Z}_{t_i}$ as 
\begin{equation}
    \bm{Z}_{t_i} \xleftarrow{} e^{-{\rm max}(0, w_i(t_i-t_{i-1}))} \bm{Z}_{t_i}.
\end{equation}

Since part of information in $\bm{X}_t$ is available, which is represented by $\bm{M}_t \odot \bm{X}_t$, this part of available state information can be used to calculate the loss function with the constructed state $\widehat{\bm{X}}_t$ by the proposed method. The loss function is defined as 
\begin{equation}
    \mathcal{L} = \frac{\sum_{t=t_0}^{T}\sum_{i=1}^{N}\sum_{j=1}^dm_t^{i,j}(\widehat{x}_t^{i,j}-x_t^{i,j})^2}{\sum_{t=t_0}^T\bm{1}^{\top}\bm{M}_t\bm{1}},
\end{equation}
where $\bm{1} \in \mathbb{R}^{N \times 1}$ is a vector of all ones. The algorithm is shown in Algorithm~(\ref{algorithm1}).  

The prediction network is trained using observable data and imputation data. The error exists between the imputation data and the ground truth inevitably. Since the imputation data is obtained by the proposed RNN-based method which is trained for one-step ahead prediction, the estimation error accumulates step by step between two observations.





The generative data close to the observable data is more accurate and is more
important for training. Therefore, instead of using the homogeneous weight in loss function, e.g., $\mathcal{L}=\frac{\sum_{i=1}^{N} ||y_i-x_i||^2}{N}$, where the weight of each term, $||y_i-x_i||^2$ is $1$, the weight $w_i$ of each term is designed based on the time interval between estimation state and the observation state as follow:

\begin{equation}\label{equ: decay loss}
    w_{t_{ij}} = \beta e^{-\zeta (t_i-t_{ij})},   
\end{equation}
where $t_i < t_{ij} <t_{i+1}$ is the time between two observable time $t_i$ and $t_{i+1}$. $\bm{X}_{t_{ij}}$ is estimated by the imputation network based on $\bm{X}_{t_{i}}$. $\zeta$ is the exponential decay constant. $\beta=\frac{\sum_{n=1}^{N}\sum_{d=1}^{D} m_{t_i}^{n,d}}{N*D}$. $\sum_{n=1}^{N}\sum_{d=1}^{D} m_{t_i}^{n,d}$ is the number of features we can observe. $N*D$ is the total number of features the state should have at time $t_i$. $\beta$ quantifies the ratio between the number of observable features and actual features. Equation~(\ref{equ: decay loss}) indicates that the estimation state close to the observable state with less lost information plays a more important role in training prediction network.

\begin{algorithm}[ht]
\caption{Graph Neural ODE with reliability and time-aware mechanism}
\begin{algorithmic}[1]
    \STATE \textbf{Input}: Time-series data $\{(\bm{X}_{t_i}, t_i)\}_{i=0, 1,2, 3,\cdots}$
    \STATE Make the initial imputation of state $\widetilde{\bm{X}}_{t_0} = \bm{M}_{t_0} \odot \bm{X}_{t_0}+\overline{\bm{M}}_{t_0} \odot \sigma(\bm{H}_{0}\bm{V}_0+\bm{b}_0)$;
    \FOR{episode $i=1, 2, 3, \cdots$}
    \STATE Estimate the hidden state between observations by Graph Neural ODE $\bm{H}'_{t_i}={\rm ODESolve} (\bm{F}_{\rm GCN}, \bm{H}_{t_{i-1}}, (t_{i-1}, t_i))$;
    \STATE Calculate the reliability matrix $\bm{U}_{t_i}$ based on $\bm{M}_{t_i}$, $\bm{X}_{t_i}$ and $\widehat{X}_{t_i}$ at each time step;
    \STATE Embedding the time-aware mechanism into the forget gate $ \bm{Z}_{t_i} \xleftarrow{} e^{-{\rm max}(0, w_i(t_i-t_{i-1}))} \bm{Z}_{t_i}$;
    \STATE Update the hidden state $\bm{H}_{t_i}$ by Graph Convolutional GRU with $\widetilde{\bm{X}}_{t_i}||\bm{U}_{t_i}||\bm{H}_{t_{i-1}}$;
    \STATE The state $\bm{X}_{t_i}$ with missing values can be imputed by $\widetilde{\bm{X}}_{t_{i+1}} = \bm{M}_{t_{i+1}} \odot \bm{X}_{t_{i+1}}+\overline{\bm{M}}_{t_i+1} \odot (\sigma(\bm{H}_{t_i}\bm{V}_s+\bm{B}_s))$ with 
    \ENDFOR
    \STATE Calculate the loss function based on observable states and prediction states $\mathcal{L} = \frac{\sum_{t=t_0}^{T}\sum_{i=1}^{N}\sum_{j=1}^dm_t^{i,j}(\widehat{x}_t^{i,j}-x_t^{i,j})^2}{\sum_{t=t_0}^T\bm{1}^{\top}\bm{M}_t\bm{1}}$
    
\end{algorithmic}\label{algorithm1}
\end{algorithm}

\section{Experiments}
We compare the proposed method in this paper to other autoregressive models such as the RNN based methods like traditional RNN, RNN $\Delta t$, RNN Decay, and Neural ODE. As the loss function $\mathcal{L}_{\rm MSE}$ captures errors throughout an entire time series we adopt this also as our evaluation metric.

We consider a simple case study of a networked dynamics system which consists of $8$ nodes and each node has a $2$-dimensional dynamics function. The dynamics of each node is 

\begin{equation} \label{equ: 2D_dynamics}
    \begin{split}
        &\dot{x}^{i,1}=-0.1*(x^{i,1})^3-2*x^{i,2}\\
        &\dot{x}^{i, 2} = x^{i,1}-0.1*x^{i, 2}  
    \end{split}
\end{equation}
The coupling function is $x^{i, 1}-x^{j, 1}$. The connection among nodes is set as 
\[
    \begin{aligned}
        &[0, 1, 2, 3, 0, 0, 5, 6]\\
        &[1, 2, 3, 4, 5, 3, 6, 7].
    \end{aligned}
\]
The data is generated by Equation~(\ref{equ: 2D_dynamics}) with irregular observation time-series points which are generated by exponential function. Part of observable states' information is randomly deleted to generate the partial observable data. The modeling of the evolution of the dynamics system is shown in Fig.~\ref{fig: predict}. The sampling data is from 0 to 10 seconds and the prediction is from 10 to 20 seconds. We randomly generate 100 points between $0$ and $10$ seconds as the ground truth and only part of the  data (e.g. $20\%, 50\%, 70\%$) is reserved as the observations. Then we randomly delete part of data at each time points as the missing  information. In total we sample 100 trajectories and keep $70\%$ as trainning set and $30\%$ as testing set. Fig.~\ref{fig: predict} shows the evolution of the system reconstructed according to the irregularly partial observed data. Table~(\ref{Tab: 1}) shows the comparison results of different methods.

\begin{table}[]\caption{Test Mean Squared Error (MSE) ($\times 10^{-2}$) on the dynamics system dataset}
\begin{tabular}{lllllll}
\hline
                & \multicolumn{3}{l}{Interpolation} & \multicolumn{3}{l}{Extrapolation} \\
                \hline
Model    &      20\% & 30\%  & 50\%   & 20\% & 30\%   &50\%      \\

RNN ($\Delta t$)     &6.72    &       3.56   &       2.87           &    8.93      &       5.73    &      3.83     \\
RNN (GRU-Decay) & 5.78          & 2.93          & 1.56          & 20.62          & 15.58          &  14.63         \\
Neural ODE      &           5.62&         2.67 &           1.21&      12.52     &     14.62      &16.67           \\
Proposed Method &          4.98 &         1.94  &           0.98&  6.72        &   3.68        &  2.87 \\ 
\hline
\end{tabular}\label{Tab: 1}
\end{table}

\section{Conclusion}
Modelling a dynamics system and predicting future states with irregularly partial observed time-series data is a challengable task.In this paper, we propose a framework which consists of an impute network and a prediction network to model the evolution of networked system by irregular sampling and partial observable time-series data. The impute network is based on temporal Graph Neural ODE consisting of Graph Neural ODE and Graph Gate Recurrent Unit with reliability and time-aware mechanism. Unlike RNN-based method with sharing RNN cell to update hidden states, the proposed method which can capture the impact of various intervals and missing state
information as well as the spatial and temporal dependencies, enabling  accurately impute temporal and spatial data. The prediction network can make prediction by learning from the imputation data. Since the quality of the imputation data generated by the impute network is heterogeneous, an exponential decay function is designed to adjust the weight of the data to calculate the loss function. This enables the sample of higher quality to play a more important role in training the prediction network. We verified our method in a networked dynamics system and compared it with other existing methods and our method can more accurately model the dynamics of the system from time-series data. However, we found that compared with other methods, using Neural ODE to calculate the hidden state is more time-consuming and sensitive to the time step when calculate the ODE. Therefore, in the future, we will explore how to accurately and efficiently estimate the hidden state between intervals. Also, it is interesting to specify the proposed method in different application fields with network structures.


%





\ifCLASSOPTIONcaptionsoff
  \newpage
\fi





\bibliographystyle{IEEEtran}
\bibliography{Bibliography}

\vfill


\end{document}